\documentclass[12pt]{article}

\usepackage[a4paper, left=3cm,top=3cm,right=3cm,bottom=3.5cm]{geometry}
\usepackage{concmath}
\usepackage[T1]{fontenc}

\usepackage[utf8]{inputenc}

\usepackage{authblk}
\usepackage[english]{babel}
\usepackage{amsmath,amsthm,amssymb}
\usepackage{dsfont}
\usepackage{mathtools}
\usepackage{subcaption}
\usepackage[color=green!40]{todonotes}

\newcommand*{\R}{\mathds{R}}

\newcommand{\mc}[1]{\mathcal{#1}}
\newcommand*{\al}[1]{{\alpha_{#1}}}
\newcommand\newsubcap[1]{\phantomcaption%
	\caption*{\figurename~\thefigure\thesubfigure: #1}}

\DeclareGraphicsExtensions{.eps,.pdf,.png,.jpg}

\usepackage{siunitx}

\numberwithin{equation}{section}
\numberwithin{figure}{section}
\numberwithin{theorem}{section}

\author{Christoph Angermann}
\author{Markus Haltmeier}

\affil{Department of Mathematics\authorcr
University of Innsbruck\authorcr
Technikerstrasse 13, 6020 Innsbruck, Austria\authorcr
 {\tt \{christoph.angermann,markus.haltmeier\}@uibk.ac.at}
 }

\author{Ruth Steiger}
\author{Sergiy Pereverzyev Jr.}
\author{Elke Gizewski}

\affil{Universit\"atsklinik f\"ur Neuroradiologie\authorcr
Medizinische Universit\"at Innsbruck\authorcr
Anichststra{\ss}e 35, 6020 Innsbruck\authorcr
{\tt neuroradiologie@i-med.ac.at}}

\title{Projection-Based 2.5D U-net Architecture  for Fast Volumetric Segmentation}

\date{August 5, 2019}

\begin{document}
\maketitle

\begin{abstract}
Convolutional neural  networks are state-of-the-art  
	for various segmentation tasks.  
	While for 2D images these networks are also computationally  efficient, 
	3D  convolutions  have  huge storage requirements and 
	require long training time.  
	To overcome this issue,   we introduce a network structure for volumetric data 
	without 3D  convolutional layers. 
	The main idea is to include maximum intensity projections from different directions  to transform the volumetric 
	data to a sequence of   images, where each image contains 
	information of the full data.
	We then apply  2D convolutions  to these projection images 
	and lift them again to volumetric data using a trainable reconstruction algorithm. 
	The proposed  network architecture has less storage requirements than network structures using 3D convolutions. For a tested binary segmentation task, it even shows better performance than the 3D U-net and can be trained much faster.
\end{abstract}

\section{Introduction}

Deep convolutional neural networks have become a powerful method for image recognition (\cite{KH16,vgg}) . In the last few years they also exceeded the state-of-the-art in providing segmentation masks for images. In \cite{JL15}, the idea of transforming VGG-nets \cite{vgg} to deep convolutional filters to obtain semantic segmentations of 2D images  came up. Based on these deep convolutional filters, the authors of   \cite{OR15} introduced a novel network architecture, the so-called U-net. With this architecture they redefined the state-of-the-art in 2D image segmentation till today. The U-net provides a powerful 2D segmentation tool for biomedical applications, since it has been demonstrated to learn highly  accurate 
ground truth masks  from  only very few training samples.

Among others,  the  fully automated generation of volumetric segmentation masks becomes increasingly important for  biomedical applications.
This task still is challenging.     One idea is to extend the U-net structure 
to volumetric data by using 3D convolutions, as has been proposed in \cite{OC16,EB17}. Essential drawbacks are the huge memory 
requirements and long training time.  Deep learning segmentation  
methods therefore are often applied to 2D slice images (compare \cite{OC16}). However, 
these slice images do not contain  information of the full 3D data which makes the segmentation task much more challenging.

\begin{figure}[htb!]
	
	\centering
	\begin{subfigure}{0.24\textwidth}
		\centering
		\includegraphics[scale=0.3]{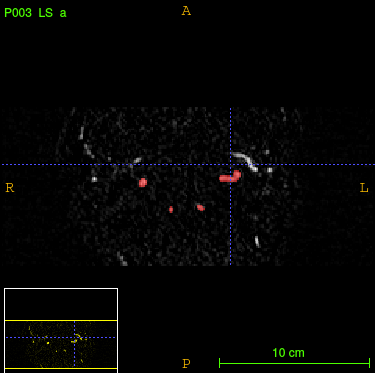}
		\subcaption{ Transversal.}
	\end{subfigure}
	\begin{subfigure}{0.24\textwidth}
		\centering
		\includegraphics[scale=0.3]{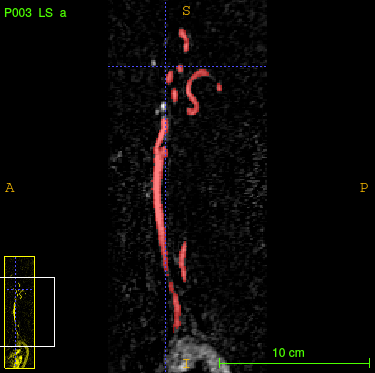}
		\subcaption{ Sagittal.}
	\end{subfigure}
	\begin{subfigure}{0.24\textwidth}
		\centering
		\includegraphics[scale=0.3]{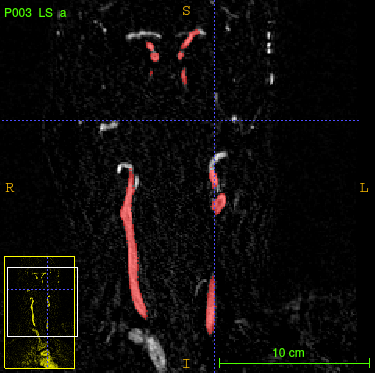}
		\subcaption{ Coronal.}
	\end{subfigure}
	\begin{subfigure}{0.24\textwidth}
		\centering
		\includegraphics[scale=0.715]{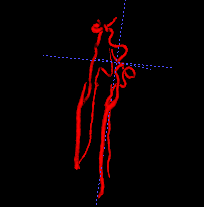}
		\subcaption{ 3D segmentation.}
	\end{subfigure}
	\caption{ In every plane (a)-(c) the blood vessels of interest are marked in red. In (d) we see the corresponding 3D segmentation mask. The segmentation was conducted with the freeware ITK-SNAP \cite{ITK}.}
	\label{fig:fig3}
\end{figure}

To address the  drawbacks  of existing  approaches, 
we  introduce  a  network structure which is able to 
generate accurate  volumetric segmentation masks of very large 
3D volumes.  The main idea is  to integrate maximum intensity projection (MIP) layers from different directions which transform the data to 2D images containing
information of the full 3D image. 
As an example, we test the network for segmenting 
blood vessels (arteries and veins)   in  magnetic resonance 
angiography (MRA) scans (Figure \ref{fig:fig3}). 

The proposed  network can be trained $15\times$ faster and requires order of magnitude  less  memory  than networks with 3D convolutions, and still 
produces more accurate results.

\section{Background}

\subsection{Volumetric segmentation of blood vessels}

We aim 
at generating volumetric binary segmentation masks.  In particular, as one targeted application, we aim   
at segmenting blood vessels 
(arteries and veins)  which assists the doctor
to detect abnormalities like stenosis or aneurysms. Furthermore, the medical sector is looking for a fully automated method to evaluate large cohorts in the future.
The Department of Neuroradiology Innsbruck has provided volumetric MRA scans of 119 different patients. The images face the arteries and veins between the brain and the chest. Fortunately, also the volumetric segmentation masks (ground truths) of these 119 patients have been provided.  These segmentation masks  have been generated by hand 
which is long hard work (Figure \ref{fig:fig3}). 

Our goal  is the fully automated generation of the 3D
segmentation masks of the blood vessels.   For that 
purpose we use deep learning and neural networks.
At the first glance, this problem may seem to be quite easy 
because we only have 
two labels (0: background, 1: blood vessel).
However, there are also arteries and veins which have label 0 which might  confuse the network since we only want to segment those vessels of interest.     Other challenges  are caused by the big size of the 
volumes ($96 \times 288 \times 224$ voxels) and by the very unbalanced distribution of the two labels (in average, 99.76 \% of all voxels indicate background).

\begin{figure}[htb!]
	\centering
	\includegraphics[scale=0.45]{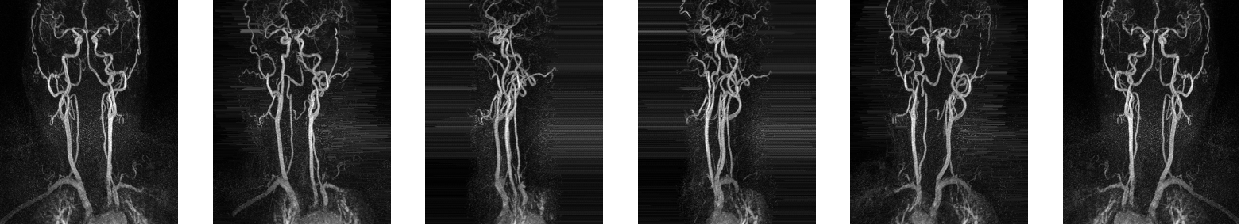}\\
	\includegraphics[scale=0.45]{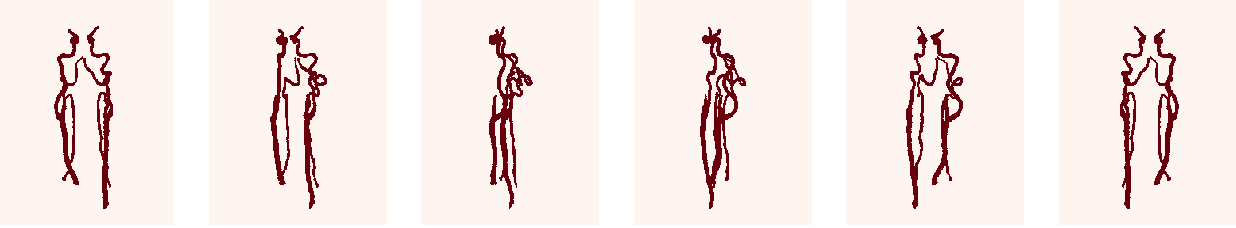}
	\caption{MIP images of a 3D MRA scan with $\alpha=36^\circ$. In the first row, we see the projections of the original scan, in the second row the corresponding projections of the ground truth.}
	\label{fig:fig5}
\end{figure}

\subsection{Segmentation of MIP images}
\label{sec:u-net}

We first solve a 2D version of  our problem. 
This can be done by applying maximum intensity projections to the 3D data  and the  corresponding 3D 
ground truths.
Using  a  rotation angle of $\alpha = 36^\circ$ around the vertical axis 
we obtain 10 MIP images out of each patient, which results in a data set 
to 1190 pairs of 2D images  and corresponding 2D segmentation masks.  Data corresponding for one patient are shown in Figure~\ref{fig:fig5}.

\begin{figure}[htb!]
	\centering
	\includegraphics[scale=0.6]{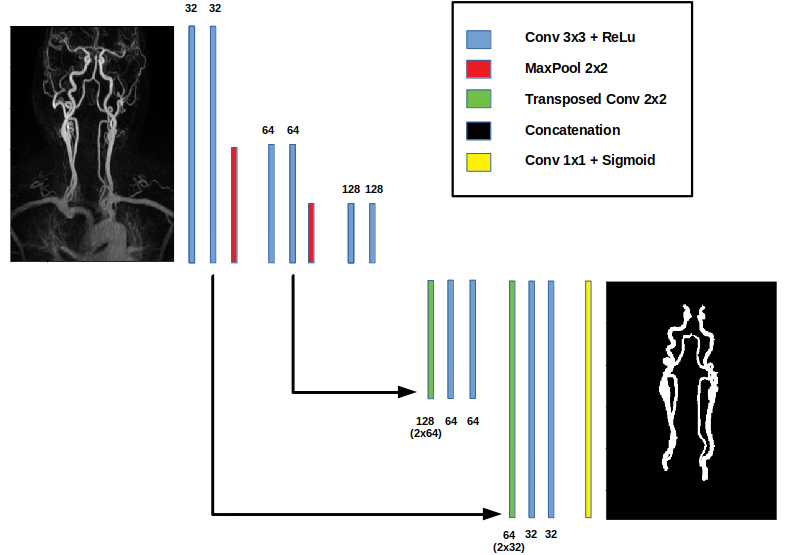}
	\caption{Visualization of the ground architecture of a 2D U-net.}
	\label{fig:fig11}
\end{figure}

The U-net used for binary segmentation  is a mapping
$\mc{U} \colon \R^{a \times b } \to [0,1]^{a \times b } 
$  which takes an image  as input and outputs 
for each pixel the probability of being a foreground pixel.    
It is  formed  by the following ingredients~\cite{OR15}:
\begin{itemize}
	\item The \textit{contracting part}: 
	It includes stacking over convolutional blocks (consisting of 2 convolutional layers) and  max-pooling 
	layers considering following properties: 
	(1)  We only use $3\times 3$ filters to hold down complexity and zero-padding to guarantee that all layer outputs have even spatial dimension.
	(2) Each max-pooling layer has stride $(2,2)$ to half the spatial sizes. We must be aware that the spatial dimensions of the input  can get divided by 2 often enough without producing any rest. This can be done by slight cropping. (3) After each pooling layer we use twice as many filters as in the previous convolutional block.
	
	\item The \textit{upsampling part}:
	To obtain similarity to the contracting part, we make use of transposed convolutions to double spatial dimension and to halve the number of filters. They are followed by convolutional blocks consisting of two convolutional layers with kernel size $3 \times 3$ after each upsampling layer (compare Figure \ref{fig:fig11}).
\end{itemize}

Every convolutional layer in this structure gets followed by a ReLu-activation-function. To link the contracting and the upsampling part, concatenation layers are used, where two images with same spatial dimension get concatenated over their channel dimension (see Figure \ref{fig:fig11}). This ensures a combination of each pixel's information with its localization. At the end, the sigmoid-activation-function is applied, which outputs  for each pixel  the probability for being a foreground pixel. 
To get the final segmentation mask,  a threshold (usually 0.5) is applied point-wise 
to the output of the U-net.\\
All networks in this paper are build with the \textit{Keras} library \cite{keras} using \textit{Tensorflow} backend \cite{tf}. Our implemented  2D U-net  has filter size 32 at the beginning and filter size 512 at the end of the contracting part. The values of the start weights are normally distributed with expectation 0 and deviation $\frac{1}{f_l}$, where $f_l$ denotes the size of the $l$-th convolutional layer. The network is trained with  the 
Dice-loss function \cite{EB17}  
\begin{equation*} 
\ell(y,\hat y) = 1 - \frac{ 2 \sum_k (y  \odot \hat y)_k }{\sum_k \hat{y}_k+\sum_k y_k},
\end{equation*}
where $\odot$ denotes  pixelwise multiplication,  
the  sums are taken over all pixel locations, 
$\hat y = \hat{f}(x) $ are the probabilities predicted 
by the U-net,  and $y$ is the related ground truth. 
The Dice-loss function measures similarity by comparing all correctly  predicted vessels pixels  with   the total number of vessels pixels 
in the prediction.

For $i, j\in\{0,1\}$   let us denote  by $p_{ij}$ the  set of all pixels of 
class  $i$  predicted to class $j$, and by  $t_i$
the number  of all pixels belonging to class $i$. 
With this notation, we  evaluate the following  metrics  
during training:
\begin{itemize}
	\item \textit{Mean Accuracy}:
	\begin{flalign*}
	&\textbf{MA}\triangleq 
	\frac{1}{2} \left(\frac{p_{00}}{t_0}+\frac{p_{11}}{t_1}\right) \,,&&
	\end{flalign*}
	
	\item \textit{Mean Intersection over Union} \cite{iou}:
	\begin{flalign*}
	&\textbf{IU}\triangleq 
	\frac{1}{2} \left(\frac{p_{00}}{t_0+p_{10}}+\frac{p_{11}}{t_1+p_{01}}\right) \,,&&
	\end{flalign*}
	
	\item \textit{Dice-coefficient} (\cite{EB17,dice}):
	\begin{flalign*}
	&\textbf{DC} \triangleq 
	\frac{2p_{11}}{2p_{11}+p_{01}+p_{10}} \,.&&
	\end{flalign*}
\end{itemize}

To guarantee that all samples have satisfying spatial dimensions, the images get symmetrically cropped a little bit.
We also make use of batch normalization layers \cite{batchnorm} before each convolutional block to speed up convergence. To handle overfitting \cite{goodfellow2016deep}, we also recommend the integration of dropout layers \cite{dropout} with dropout rate 0.5 in the deepest convolutional block and dropout rate 0.2 in the second deepest blocks.  For training,  Adam-optimizer \cite{adam} is used with learning rate 0.001 in combination with learning-rate-scheduling, i.e. if the validation loss does not decrease within 3 epochs the learning rate gets reduced by the factor 0.5. Furthermore, if the network shows no improvement for 5 epochs the training process gets stopped (early stopping) and the weights of the best epoch in terms of validation loss get restored. We use a (70, 15, 15) split in training, validation, and evaluation data and a threshold of 0.5 for the construction of the segmentation masks. Training the U-net on    
\textit{NVIDIA GeForce RTX 2080}  GPU with a minibatch size of 6 yields the following 
results:
Dice-loss of 0.088,
mean accuracy of 95.7\%,
mean IU of 91.6\%,
and Dice-coefficient of 91.3\%. In average, training the 2D U-net lasts 809 seconds, the application only 0.013 seconds. During training, the 2D U-net allocates a memory space of 1.7 gigabytes.

\subsection{Segmentation with the  3D U-net}

Now we aim at generating binary segmentation masks of sparse volumetric data using a 3D version of the prior introduced U-net. The resulting 3D U-net follows the same structure as in \ref{sec:u-net}, the only difference is the usage of 3D convolutions and 3D pooling layers. 
For the 3D U-net we have to take special care about overfitting \cite{goodfellow2016deep} and about memory space. Therefore, for the  3D U-net we have  chosen
filter size 4 at the beginning and filter size 16 at the end of the contracting part. Also the use of high dropout rates \cite{dropout} (0.5 in the deepest convolutional block and 0.4 in the second deepest blocks) is necessary to ensure an efficient training process. Due to the huge size of our training samples ($96 \times 288 \times 224$ voxels), we train the network on batch, i.e. with minibatch size 1. During training the 3D U-net allocates more than 8 gigabyte memory space and therefore is not manageable any more by our GPU. Therefore, the training process on a \textit{AMD Ryzen 7 1700X} eight-core processor takes   
in average 969 minutes. Since the number of 3D samples is only 119, we conducted 5 training-runs with random choice of training, validation and evaluation data. 
Using the 3D U-net  we obtained in average following results:
Dice-loss of 0.254, mean accuracy of 87.3\%, mean IU of 80.5\% and Dice-coefficient of 74.8\%.

Although the 3D U-net  demonstrates   high precision   in  our application (see Figures \ref{fig:fig2},\ref{fig:fig1}), we are not satisfied with the long training time. In addition 
to it, we are very limited in the choice of convolutional layers and the corresponding number of filters due to the huge size of the input data. So it is hardly possible to conduct volumetric segmentation for even larger biomedical scans without using cropping or sliding-window techniques.

\section{Projection-Based 2.5D U-net}

 As mentioned in the introduction, the naive approach for accelerating volumetric segmentation and reducing memory requirements is to process each of the 96 slice images independently through a 2D-network (compare \cite{OC16}). However, this causes the loss of connection between the slice images. For our targeted application, applying the 2D U-net out of \ref{sec:u-net} to each slice image of the 3D MRA scans independently yields following disappointing results: Dice-loss of 0.849,
mean accuracy of 54.5\%,
mean IU of 54.3\%,
and Dice-coefficient of 15.1\%. Therefore we are looking for an alternative approach.

\begin{figure}[htb!]
	\centering	
	\includegraphics[scale=0.7]{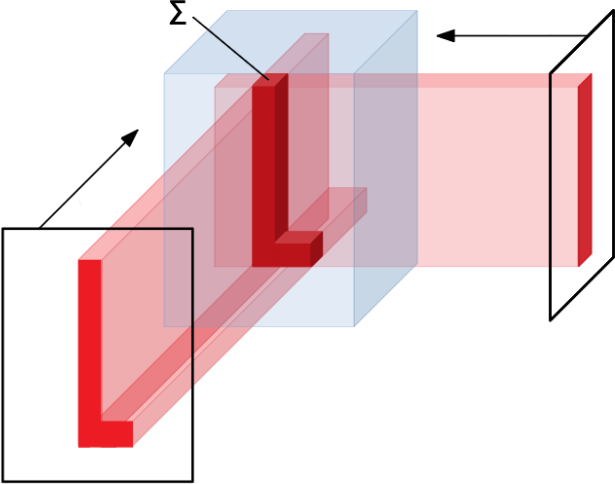}
	\newsubcap{ \textbf{Reconstruction operator} $\mc{R}_2$: Voxel value is defined as the sum over the corresponding 2D values, here illustrated for 2 MIP images with directions $\{0^\circ ,90^\circ\}$.}
	\label{fig:fig18}
\end{figure}

\subsection{Proposed 2.5D U-net architecture}
As we have seen in \ref{sec:u-net}, the 2D U-net does very well on the MIP images. Recall that a network for binary volumetric segmentation is a function $\mathcal{N}: \mathbb{R}^{a \times b \times c} \to [0,1]^{a \times b \times c }$ that maps the 3D scan to the probabilities that a voxel corresponds to the desired class. For a 3D input $x$, the proposed 2.5D U-net takes the form
\begin{align}\label{eq1}
\mc{N}({x})=\mc{T}\circ\mc{R}_p\circ\mc{F}_p\circ
\begin{bmatrix}
\mc{U} \circ \mc{M}_\al{1} ({x})
\\ \vdots \\ 
\mc{U} \circ \mc{M}_\al{p} ({x})
\end{bmatrix},
\end{align}
where

\begin{itemize}
	\item $\mc{M}_i: \mathbb{R}^{a\times b\times c}\to \mathbb{R}^{b\times c}$ are MIP images for different projection directions $\al{1},\ldots,\al{p}$,
	\item $\mc{U}:\mathbb{R}^{b\times c}\to [0,1]^{b\times c}$ is the same 2D U-net as in \ref{sec:u-net} producing probabilities,
	\item $\mc{F}_p:([0,1]^{b\times c})^p\to (\mathbb{R}^{b\times c})^p$ is a learnable filtration,
	\item $\mc{R}_p:(\mathbb{R}^{b\times c})^p \to \mathbb{R}^{a\times b\times c}$ is a reconstruction operator using $p$ \textit{linear backprojections} as shown in Figure \ref{fig:fig18},
	\item $\mc{T}:\mathbb{R}^{a\times b\times c} \to [0,1]^{a\times b\times c}$ is a fine-tuning operator (average pooling followed by a learnable \textit{shift}-operator followed by the sigmoid-activation-function).
\end{itemize}

\begin{figure}[htb!]
\begin{subfigure}{0.45\textwidth}
	\centering
	\includegraphics[scale=0.6]{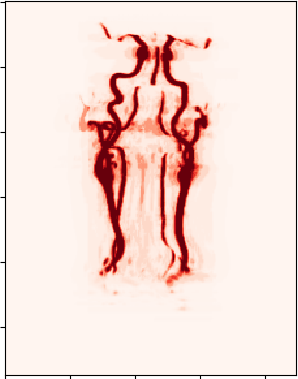}			
	\newsubcap{Network's output (before threshold) without filtration.}
	\label{fig:fig19}
\end{subfigure}\qquad
\begin{subfigure}{0.45\textwidth}
	\centering
	\includegraphics[scale=0.6]{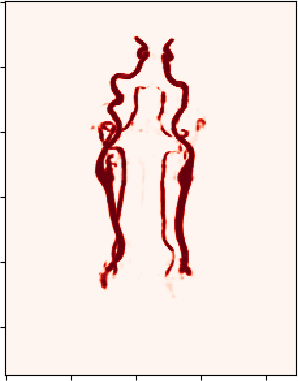}
	\newsubcap{Network's output (before threshold) with filtration.}
	\label{fig:fig199}
	
\end{subfigure}
\end{figure}

The backprojection operator $\mc{R}_p$ causes a kind of shroud (Figure \ref{fig:fig19}), so we have to think about a filtrated backprojection. Therefore, we apply a convolutional layer $\mc{F}_p$ before backprojection. Using  $1\times 3$ filters, which get adapted during training for each projection direction  $\alpha_1,\ldots,\alpha_p$ individually, leads to a more satisfying result (Figure \ref{fig:fig199}).

For the fine-tuning operator $\mc{T}$ we use average pooling with pool-size (2, 2, 2). This is followed by a learnable shift-operator, which shifts the pooled data by an  adjusted parameter since the decision boundaries have been changed by $\mc{R}_p$. This ensures, that the application of the sigmoid function delivers accurate probabilities.

For our targeted application, again we only process one 3D sample through the network per iteration (minibatch size 1). The start weights of the convolutional part $\mc{U}$ in  \ref{eq1} are initialized in the same way as in Section\ref{sec:u-net}. The parameters of $\mc{F}_p$ and $\mc{T}$ are initialized empirically.

\begin{figure}[htb!]
	\centering
	\includegraphics[scale=0.35]{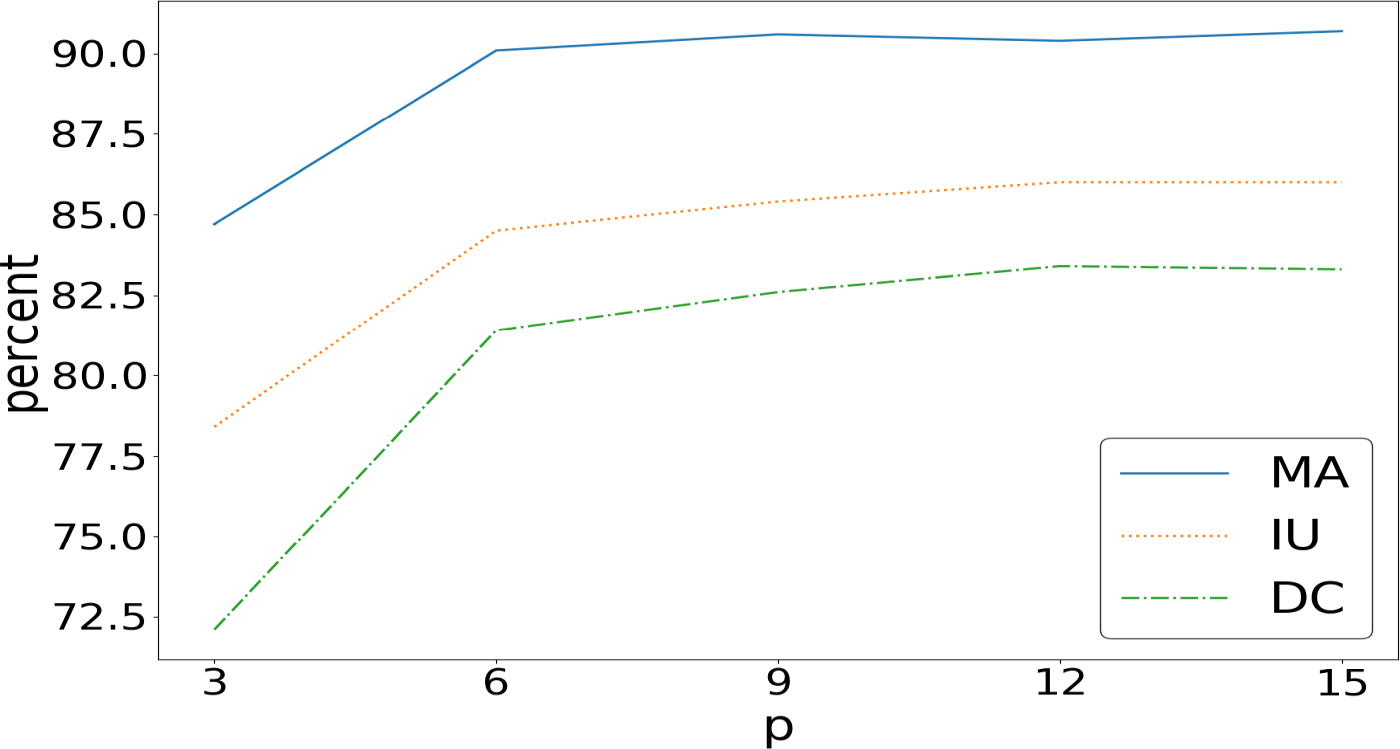}
	
	\caption{Performance of the 2.5D U-net for different number $p$ of projection directions .}
	\label{fig:fig30}
\end{figure}

For the amount of the projection directions we choose equidistant angles $\Theta=\{k\times\frac{180}{p}\mid k=0,\ldots,p-1\}$ to ensure we obtain at most different information of the 3D data for different projection directions. This causes the task of finding the best value for $p$ in \ref{eq1}. Therefore we trained the proposed network $\mc{N}$ for different values for $p$ and compared performance in terms of the evaluation metrics (Figure \ref{fig:fig30}).

Looking at Figure \ref{fig:fig30}, we observe that $\Theta =\{k\times \frac{180}{12}\mid k=0,\ldots,11\}$ seems to be a good choice for the amount of projection directions. We have conducted 5 training runs with random choice of training, validation and evaluation data and obtained in average following results:
Dice-loss of 0.201, mean accuracy of 91.6 \%, mean IU of 
86.1 \% and Dice-coefficient of 83.7 \%.

\begin{figure}[htb!]
		\centering
		\includegraphics[scale=0.41]{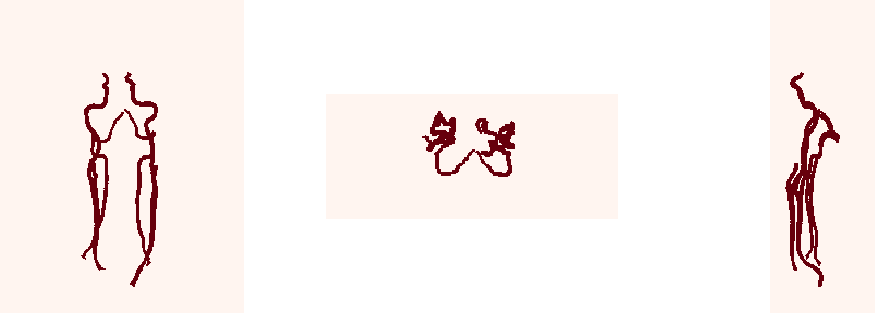}\\
		\includegraphics[scale=0.41]{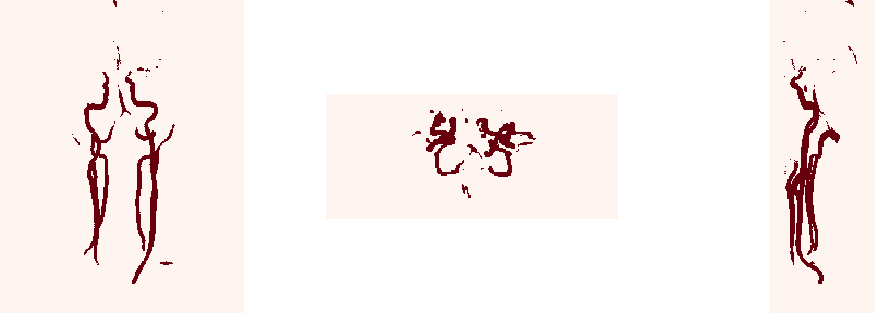}\\
		\includegraphics[scale=0.33]{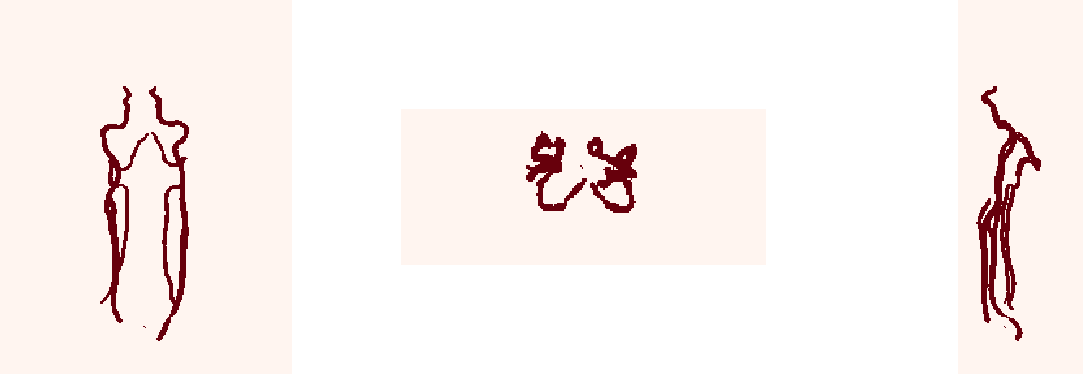}
		\newsubcap{Comparison between ground truth (first row), segmentation generated by 3D U-net (second row) and  segmentation generated by 2.5D U-net (third row).}
		\label{fig:fig2}
\end{figure}

	\begin{figure}[htb!]
		\centering
		\includegraphics[scale=0.27]{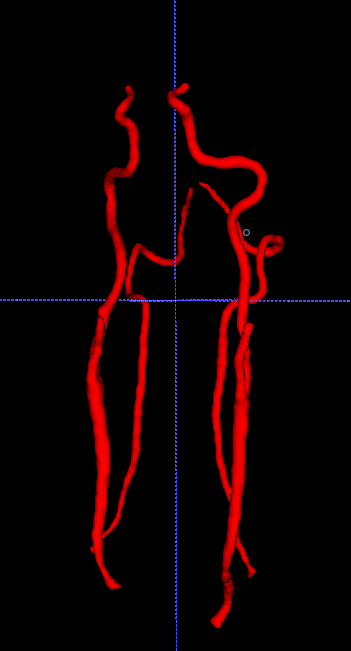}\hfil
		\includegraphics[scale=0.27]{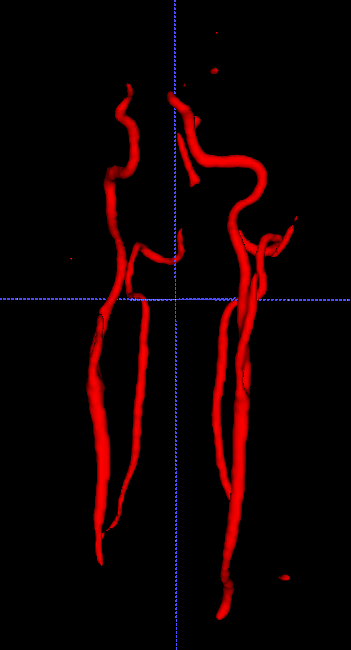}\hfil
		\includegraphics[scale=0.27]{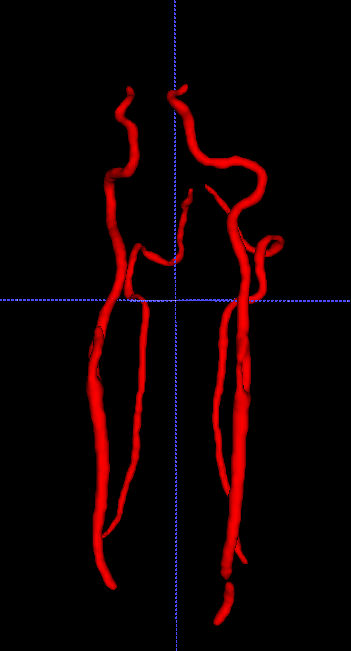}
		\newsubcap{3D segmentation mask generated by hand (left), by 3D U-net (middle) and by  2.5D U-net (right).}
		\label{fig:fig1}
\end{figure}

As we can see, for the volumetric segmentation of MRA scans the proposed 2.5D U-net clearly outperforms 3D U-net in terms of evaluation metrics. Furthermore, adjusting the weights of 2.5D U-net only takes in average 3914 seconds. With 11.37 seconds, the application time increased due to the construction of the MIP images. During training, the 2.5D U-net allocates a memory space of 3.7 gigabytes.\\
Further tasks would be to investigate if applying data augmentation techniques to the 3D samples increases accuracy of the 3D U-net. Considering our application, data deformation could cause problems due to the fact, that the orientation of the vessels has huge impact to the network's prediction. We will investigate that in the future.

\begin{table}[htb!]
\centering
	\caption{ Summarization of the evaluation results for the naive 2D U-net slice-per-slice approach, the 3D U-net and the proposed 2.5D U-net.}
	\begin{tabular}{|r || r|r |r |r |r|r|r|}
		\hline
		\textbf{Network}&\textbf{loss}&\textbf{MA in \%}&\textbf{IU in \%}&\textbf{DC in \%} \\
		\hline
		2D U-net&0.849&54.5&54.3&15.1\\ \hline
		3D U-net&0.254&87.3&80.5&74.8\\ \hline
		
		2.5D U-net &0.201&91.6&86.1&83.7\\ \hline
	\end{tabular}
	\label{tab1}
\end{table}

\begin{table}[htb!]
\centering
	\caption{ Summarization of time and storage observations for the naive 2D U-net slice-per-slice approach, the 3D U-net and the proposed 2.5D U-net..}
	\begin{tabular}{|r || r|r |r |r |r|r|r|}
		\hline
		\textbf{Network}&\textbf{Weights}&\textbf{Train}&
		\textbf{Appl.}&\textbf{Mem.} \\
		\hline
		2D U-net&$8.6\times 10^6$&809 sec.&1.83 sec.&1.7 Gb\\ \hline
		3D U-net&$2.5\times 10^4$&58140 sec.&5.28 sec.&$>8$ Gb\\ \hline
		2.5D U-net &$8.6\times 10^6$&3914 sec.&11.37 sec.&3.7 Gb\\ \hline
	\end{tabular}
	\label{tab2}
\end{table}
\section{Conclusion}
In this paper we proposed a new projection-based 2.5D U-net 
structure for fast volumetric segmentation.  The construction of volumetric segmentation masks with the help of a 3D U-net delivers very satisfying results, but the long training time and the big need of memory space are hardly sustainable. The 2.5D U-net using 12 deterministic projection directions is able to conduct 3D segmentation of very big biomedical 3D scans as reliable as the 3D U-net and can be trained much faster without any concern about memory space.  For our targeted application, the 2.5D U-net enables the generation of 3D segmentations in a storage efficient way more accurate than other approaches using 3D convolutions and can be trained almost $15\times$ faster. All numerical results considering the evaluation metrics are displayed in Table \ref{tab1}. Average training time, application time and storage requirements for each network are summarized in Table \ref{tab2}. In the current implementation, we only use MIP images for deterministic projection directions. In future work, we will investigate the use of random projection directions for network training. This could provide the possibility to use all available information from each projection direction for the construction of 3D segmentation masks. Also the conduction of comparative studies will be a future task with the aim to research, if the 2.5D U-net also increases accuracy in other applications compared to 3D convolutions.


\begin{thebibliography}{10}
\providecommand{\url}[1]{#1}
\csname url@samestyle\endcsname
\providecommand{\newblock}{\relax}
\providecommand{\bibinfo}[2]{#2}
\providecommand{\BIBentrySTDinterwordspacing}{\spaceskip=0pt\relax}
\providecommand{\BIBentryALTinterwordstretchfactor}{4}
\providecommand{\BIBentryALTinterwordspacing}{\spaceskip=\fontdimen2\font plus
\BIBentryALTinterwordstretchfactor\fontdimen3\font minus
  \fontdimen4\font\relax}
\providecommand{\BIBforeignlanguage}[2]{{%
\expandafter\ifx\csname l@#1\endcsname\relax
\typeout{** WARNING: IEEEtran.bst: No hyphenation pattern has been}%
\typeout{** loaded for the language `#1'. Using the pattern for}%
\typeout{** the default language instead.}%
\else
\language=\csname l@#1\endcsname
\fi
#2}}
\providecommand{\BIBdecl}{\relax}
\BIBdecl

\bibitem{KH16}
K.~He, X.~Zhang, S.~Ren, and J.~Sun, ``Deep residual learning for image
  recognition,'' in \emph{Proceedings of the IEEE conference on computer vision
  and pattern recognition}, 2016, pp. 770--778.

\bibitem{vgg}
K.~Simonyan and A.~Zisserman, ``Very deep convolutional networks for
  large-scale image recognition,'' \emph{arXiv preprint arXiv:1409.1556}, 2014.

\bibitem{JL15}
J.~Long, E.~Shelhamer, and T.~Darrell, ``Fully convolutional networks for
  semantic segmentation,'' in \emph{Proceedings of the IEEE conference on
  computer vision and pattern recognition}, 2015, pp. 3431--3440.

\bibitem{OR15}
O.~Ronneberger, P.~Fischer, and T.~Brox, ``U-net: Convolutional networks for
  biomedical image segmentation,'' in \emph{International Conference on Medical
  image computing and computer-assisted intervention}.\hskip 1em plus 0.5em
  minus 0.4em\relax Springer, 2015, pp. 234--241.

\bibitem{OC16}
{\"O}.~{\c{C}}i{\c{c}}ek, A.~Abdulkadir, S.~S. Lienkamp, T.~Brox, and
  O.~Ronneberger, ``3D {U}-net: learning dense volumetric segmentation from
  sparse annotation,'' in \emph{International conference on medical image
  computing and computer-assisted intervention}.\hskip 1em plus 0.5em minus
  0.4em\relax Springer, 2016, pp. 424--432.

\bibitem{EB17}
B.~Erden, N.~Gamboa, and S.~Wood, ``3D convolutional neural network for brain
  tumor segmentation,'' 2018.

\bibitem{ITK}
P.~A. Yushkevich, J.~Piven, H.~Cody~Hazlett, R.~Gimpel~Smith, S.~Ho, J.~C. Gee,
  and G.~Gerig, ``User-guided 3D active contour segmentation of anatomical
  structures: Significantly improved efficiency and reliability,''
  \emph{Neuroimage}, vol.~31, no.~3, pp. 1116--1128, 2006.

\bibitem{keras}
F.~Chollet \emph{et~al.}, ``Keras,'' \url{https://keras.io}, 2015.

\bibitem{tf}
\BIBentryALTinterwordspacing
M.~Abadi, A.~Agarwal, P.~Barham, E.~Brevdo, Z.~Chen, C.~Citro, G.~S. Corrado,
  A.~Davis, J.~Dean, M.~Devin, S.~Ghemawat, I.~Goodfellow, A.~Harp, G.~Irving,
  M.~Isard, Y.~Jia, R.~Jozefowicz, L.~Kaiser, M.~Kudlur, J.~Levenberg,
  D.~Man\'{e}, R.~Monga, S.~Moore, D.~Murray, C.~Olah, M.~Schuster, J.~Shlens,
  B.~Steiner, I.~Sutskever, K.~Talwar, P.~Tucker, V.~Vanhoucke, V.~Vasudevan,
  F.~Vi\'{e}gas, O.~Vinyals, P.~Warden, M.~Wattenberg, M.~Wicke, Y.~Yu, and
  X.~Zheng, ``{TensorFlow}: Large-scale machine learning on heterogeneous
  systems,'' 2015, software available from tensorflow.org. [Online]. Available:
  \url{http://tensorflow.org/}
\BIBentrySTDinterwordspacing

\bibitem{iou}
A.~Rosebrock, ``Intersection over union (IOU) for object detection,''
  \url{https://www.pyimagesearch.com/2016/11/07/intersection-over-union-iou-for-object-detection/},
  2016, last accessed: 2019-04-29.

\bibitem{dice}
L.~R. Dice, ``Measures of the amount of ecologic association between species,''
  \emph{Ecology}, vol.~26, no.~3, pp. 297--302, 1945.

\bibitem{batchnorm}
S.~Ioffe and C.~Szegedy, ``Batch normalization: Accelerating deep network
  training by reducing internal covariate shift,'' \emph{arXiv preprint
  arXiv:1502.03167}, 2015.

\bibitem{goodfellow2016deep}
I.~Goodfellow, Y.~Bengio, and A.~Courville, \emph{Deep learning}. MIT press, 2016.

\bibitem{dropout}
N.~Srivastava, G.~Hinton, A.~Krizhevsky, I.~Sutskever, and R.~Salakhutdinov,
  ``Dropout: a simple way to prevent neural networks from overfitting,''
  \emph{The Journal of Machine Learning Research}, vol.~15, no.~1, pp.
  1929--1958, 2014.

\bibitem{adam}
J.~Brownlee, ``Gentle introduction to the adam optimization algorithm for deep learning,''
  \url{https://machinelearningmastery.com/adam-optimization-algorithm-for-deep-learning/},
  2017, last accessed: 2019-01-29.

\end{thebibliography}
\end{document}